\documentclass[10pt,twocolumn,letterpaper]{article}

\usepackage{cvpr}              % To produce the CAMERA-READY version
\usepackage{graphicx}
\usepackage{amsmath}
\usepackage{amssymb}
\usepackage{booktabs}
\usepackage{bm}
\usepackage{booktabs}
\usepackage[pagebackref,breaklinks,colorlinks]{hyperref}

\usepackage[capitalize]{cleveref}
\crefname{section}{Sec.}{Secs.}
\Crefname{section}{Section}{Sections}
\Crefname{table}{Table}{Tables}
\crefname{table}{Tab.}{Tabs.}

\def\cvprPaperID{*****} % *** Enter the CVPR Paper ID here
\def\confName{CVPR}
\def\confYear{2023}

\begin{document}

% \title{\LaTeX\ Author Guidelines for \confName~Proceedings}
% \title{2nd Place Solution to Meta AI Video Similarity Challenge on Descriptor Track for Video Copy Detection}
\title{A Similarity Alignment Model for Video Copy Segment Matching}

 \author{Zhenhua Liu\textsuperscript{\rm 1}\footnotemark[1],Feipeng Ma\textsuperscript{\rm 1,2}\thanks{Equal contribution.}, Tianyi Wang\textsuperscript{\rm 1}\footnotemark[1], Fengyun Rao\textsuperscript{\rm 1}\thanks{Corresponding author.}\\
 \textsuperscript{\rm 1}WeChat of Tencent, 
 \textsuperscript{\rm 2}University of Science and Technology of China  \\
 {\tt\small mafp@mail.ustc.edu.cn, \{edinliu, tyewang, fengyunrao\}@tencent.com}}

% \footnotetext[1]{Equal contribution.}

\maketitle

% \footnotetext[1]{Equal contribution.}
% \footnotetext[2]{Corresponding author.}

%%%%%%%%% ABSTRACT
\begin{abstract}
With the development of multimedia technology, Video Copy Detection has been a crucial problem for social media platforms. Meta AI hold Video Similarity Challenge on CVPR 2023 to push the technology forward. In this report, we share our winner solutions on Matching Track. We propose a Similarity Alignment Model(SAM) for video copy segment matching. Our SAM exhibits superior performance compared to other competitors, with a 0.108 / 0.144 absolute improvement over the second-place competitor in Phase 1 / Phase 2.
Code is available at \url{https://github.com/FeipengMa6/VSC22-Submission/tree/main/VSC22-Matching-Track-1st}.
   
\end{abstract}

%%%%%%%%% BODY TEXT
\section{Introduction}
\label{sec:intro}
In the past decade, the development of information technology has led to a shift in the main carrier of information from text to images and then to videos. Moreover, with the rise of User-generated Content (UGC), the producer of information has shifted from Occupationally-generated Content (OGC) to UGC. As a result, a large number of videos have emerged on social media platforms and have been widely shared, leading to the increasingly important and challenging problems of video copyright protection. 
The video copy detection task can always be divided into two parts: the descriptor task, which is used to recall similar videos, and the matching task, which is used to locate the copied segment. In this report, we summarize our work on the Matching Track of the Meta AI Video Similarity Challenge.

\begin{figure}[]
    \centering
    \includegraphics[width=0.48\textwidth]{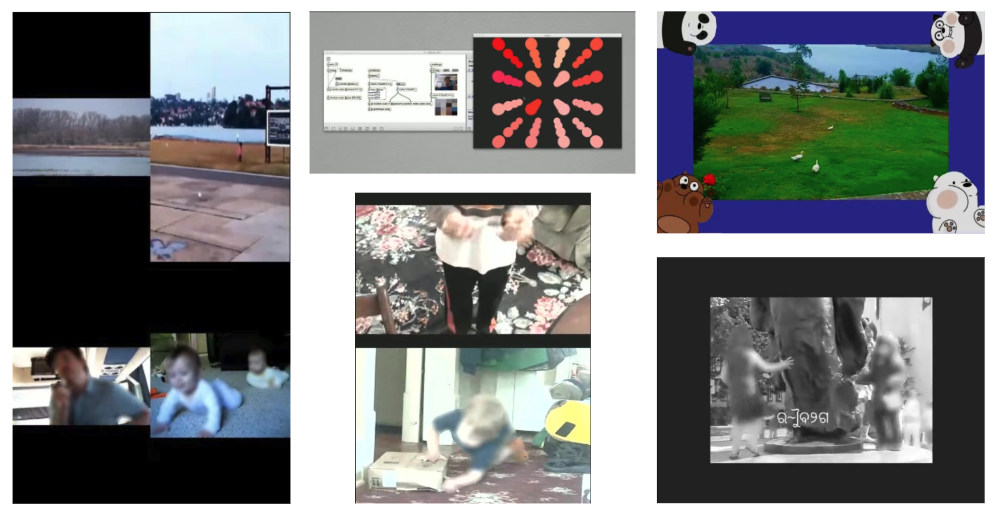}
    \caption{Typical edits in stacked frames and extra edges. A plenty of video contains 2-4 scenes, as showed in the first two columns. And there is also extra edges showed in the last column.}
    \label{fig:split}
\end{figure}

For the matching task, two important problems arise. The first is deciding which feature to use: the embedding or the similarity matrix. The second is determining how to match the video copy segments. We first consider the feature. It is reasonable choice to share features among the matching task and the descriptor task, as two task are highly correlated. Extracting features independently for each task would require double the computing resources, so sharing features can reduce the computation cost. As to the model input, the advantage of using embedding for the matching task is that it contains more information and can be used for further tuning\cite{he2022transvcl} . However, the drawback is that the matching model must be changed when the descriptor model is changed, as two different embedding models may have little correlation with each other. While the similarity matrix is more robust as changes of embedding model does not change the characteristic of similarity matrix, and even the use of different embedding models can expand the limited annotations. Finally, we choose the similarity matrix as model input.

\begin{figure*}[]
    \centering
    \includegraphics[width=1\textwidth]{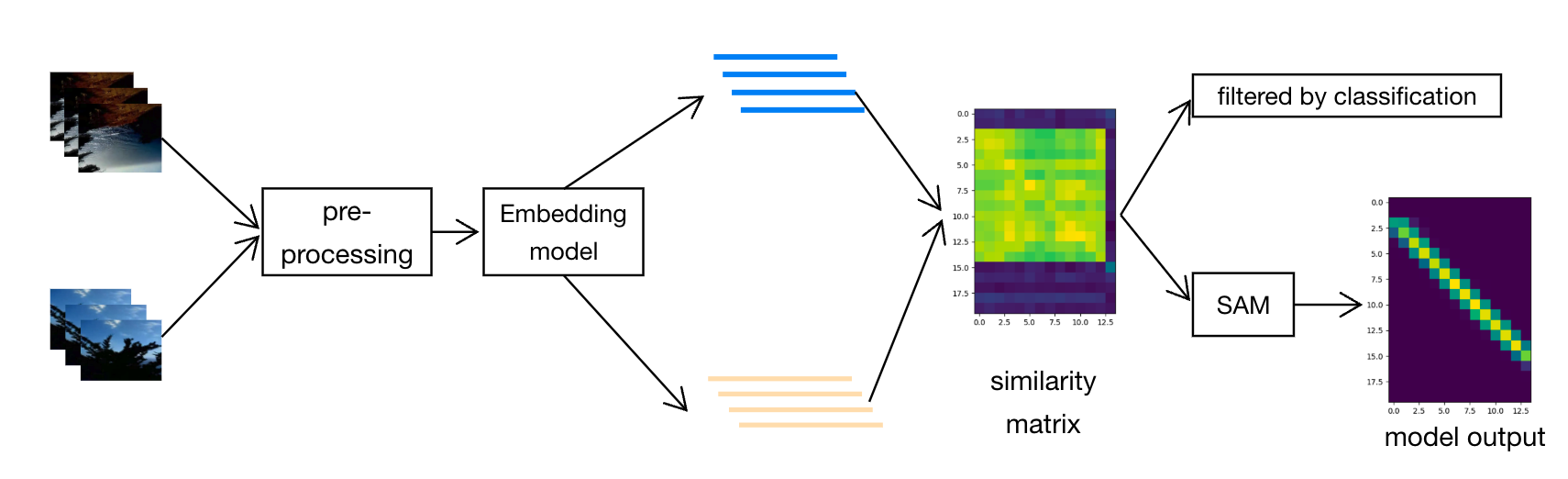}
    \caption{Our Pipeline: Preprocess input video by extracting frames, splitting scenes, and removing edges. Use embedding model(s) to generate embeddings. Generate similarity matrix for (query, reference) videos, filter negative recalls using a small classification model. Use SAM model to output noiseless matching score matrix.}
    \label{fig:pipeline}
\end{figure*}

As to the video copy segments, we found several academic approaches. First is the Temporal Network(TN)\cite{tan2009scalable}, a graph-based method takes matched frames as nodes and similarities between frames as weights of links to construct a network. This is also the baseline, but we found this method is hard to modify and optimize. Similarity Pattern Detection (SPD)\cite{jiang2021learning} adopts detection method to direct output the result. We attempted to use this method, but encountered difficulties in optimizing the model with limited annotations. We also explored TransVCL\cite{he2022transvcl}, but ultimately decided to against it. Because this method relies on frame embedding as input, which does not align with our project's objectives. Most importantly, we discovered that the primary challenge here is not simply outputting the results in an end-to-end way, but rather obtaining a cleaner matching relationship in comparison to the raw similarity matrix input. To address this, our SAM was specifically designed to take a similarity matrix as input and output a score matrix with the same resolution, with significantly improved matching relationships.

\section{Method}
In this section, we will introduce the whole pipeline we used to develop our result. As shown in figure \ref{fig:pipeline}. Our pipeline include the preprocessing, embedding extraction, similarity matrix filtering, and the SAM model processing.

\subsection{Preprocessing}

\begin{figure}[]
	\centering
	\includegraphics[width=0.5\textwidth]{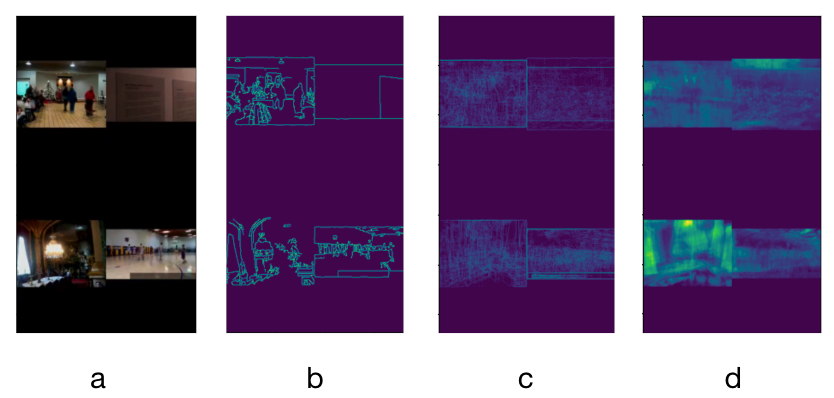}
	\caption{The frame processing details. $a$ is one frame of a query video. $b$ is the Canny \cite{canny1986computational} edge result, which is noisy to locate all edges of the stacked video. $c$ is the average Canny result of multiple frames, the edge is more clear than single frame.  $d$ is standard deviation of frame pixels values. Our frame preprocessing is base on feature $c$ and $d$ }		
	\label{fig:split_v2}
\end{figure}

Video frames were extracted at one frame per second, but many videos contained multiple scenes in one frame or extra edges. Canny \cite{canny1986computational} edge detection and frame pixel standard deviation feature was used to address this issue. First,  we average the edge detection results from multiple frames to get more robust edges. Then we use frame pixel standard deviation feature to find potential background. The whole processing is done recursively by: 1) A split images function divides a video into segments based on: $a$) the vertical or horizontal edges that extend across the frame; $b$) a low pixel standard deviation zone that split video vertically or horizontally. 2) A edge erase function that remove low variant parts of videos. With a video input, it stops when the processing parts doesn't change in size or has too low resolution. Figure \ref{fig:split_v2} shows the feature we used in preprocessing. Figure \ref{fig:split}. reveals the typical edits in stacked frames and extra edges. 

\begin{figure*}[]
	\centering
	\includegraphics[width=0.85\textwidth]{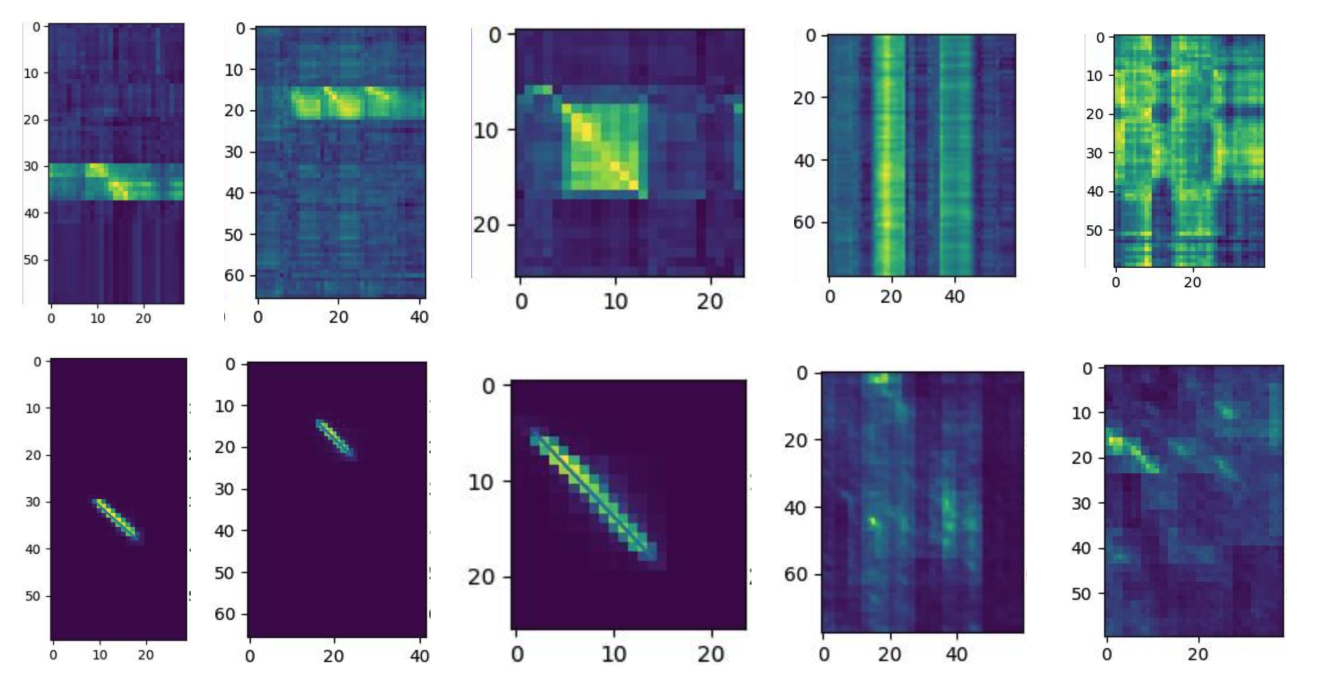}
	\caption{The input similarity matrix is compared with the SAM output matching score. The first row is the input similarity matrix, the second row is the SAM matching score. The first three column are sample successfully matched, the last two column are samples that not recognize copy result.}
	\label{fig:multiple_scenes}
\end{figure*}

\subsection{Embedding Model}

As previously mentioned, our method utilizes the embedding model from the Descriptor track, using the similarity matrix as the input for matching task. This allows our model to be resilient to changes in the embedding, and multiple embeddings of the same (query, reference) pair can be utilized as a form of data augmentation.

To maximize efficiency and recall, a recall process is employed to identify potential copied video pairs. This process only considers the similarity of descriptor features and utilizes a low recall threshold, resulting in a high number of potential matches being identified but is not copied at all. We utilized a classification process that takes a similarity matrix as input and outputs the probability that a video pair is duplicated. Our chosen classification model is the MobileNet-v3\cite{howard2019searching}, pre-trained by Image-net. This approach successfully removed 95\% of the recalled samples without impacting overall performance.

\subsection{Similarity Alignment Model}

We choose the similarity matrix as model input. The query/reference longer than 128 seconds is truncated, while the query/reference video shorter than 128 are padding to 128 with zero embedding.  So our SAM model takes (128, 128) resolution similarity matrix as input.  For query videos that has been splited into multiple frames in preprocessing. We choose the segment with max top matching similarity as target pair. Other segments with lower matching similarity are discarded. 

To build a model that output matching relationship, one possible approach is to use models such as key point detection or semantic segmentation. The key idea here is that the model should both learn global information which helps to recognize the real matching parts and the local information for percise detection. So we choose the high resolution network(HRnet-w18)\cite{SunXLW19} as our backbone.  There are two major changes to the model: 1) Our model outputs the same resolution feature maps as the input. We do it by setting the first two convolution stride to (1,1). 2) The target output has been changed to a heat-map generated by annotations to accurately reflect the real matching relationship. Figure \ref{fig:multiple_scenes}. shows some model detection result and the post-processing result.

\subsection{Postprocessing}

The SAM model outputs a matching relationship matrix, but post-processing is required before submitting the final result. This involves: 1) using a filter threshold to remove false positive matches, 2) identifying multiple detections with the Connected Components algorithm, 3) and detecting the matching relation with RANSAC \cite{fischler1981random} regression, which is effective for detecting linear relationships in video copies. The final submit score $s$ is an ensemble of SAM score predictions:
	
	$c = max(1/coef, coef)$
	
	$s = mean(score) - \alpha * std(score) - abs(c - 1) / 10 $

	Where $coef$ is the slope of RANSAC regression. $score$ is the matched score list for points which is first filter by score threshold $t$, then filter by RANSAC regression inner points. $\alpha$ is the weights for score variance penalty. For the final submission, we ensemble results with parameter ($t$, $\alpha$) equal to (0.35, 0.5), (0.1, 1.25), (0.001, 2).

\section{Experiments}

The SAM model was trained using 1 A100 GPU. After applying the classification filter, there were 10,000 pairs of samples, and their labels were generated through annotations. The resolution of the similarity matrix used for training the SAM model is 128x128, and the batch size is 64. As mentioned earlier, four embedding models from the descriptor track were used to generate the similarity matrix. Therefore, the training sample size for each round is approximately 20,000 pairs. It takes around 3 hours to finished training.

To generate the final submission, the two cross-validation models were ensembled by averaging their predicted scores. 
As to different embedding model, the SAM is directly evaluated on their PCA ensembled feature similarity matrix. 
On Matching Track,we got the first place on both phases with a 0.108 $\mu AP$ / 0.144 $\mu AP$
 absolute improvement over the second-place competitor  in Phase 1 / Phase 2.

\begin{table}[htbp]
\centering
\scalebox{0.9}{
\begin{tabular}{l|cc}
    \toprule
    User or teams & Phase 1 $\mu$AP  & Phase 2 $\mu$AP \\
    \hline
    \textbf{do something more(Ours)} & \textbf{0.9290} & \textbf{0.9153} \\ 
    CompetitionSecond & 0.8206 & 0.7711 \\ 
    cvl-matching & 0.7727 & 0.7036 \\ 
    Zihao & 0.5687&  - \\ 
    \bottomrule
\end{tabular}
}
\caption{Leaderboard results on Matching Track.}
\label{tab:matching}
\end{table}

\section{Conclusion}
 This report introduces the SAM for video copy segment matching. By modify the structure and target of high resolution network, Our model takes similarity matching matrix as input, output high quality video segment matching score. Based on this model, We get 1$^{st}$  rank on VSC 2022 Matching Track.

%%%%%%%%% REFERENCES
{\small
\bibliographystyle{ieee_fullname}
\bibliography{egbib}
}

\end{document}